\documentclass[%
 reprint,
 amsmath,amssymb,
 aps,
]{revtex4-1}
\usepackage{xcolor}
\usepackage{physics}
\usepackage{ amssymb }

\usepackage{booktabs}

\usepackage{graphicx}
\usepackage{dcolumn}
\usepackage{bm}
\usepackage{color}
\usepackage{comment}

\newcommand{\mat}[1]{\textbf{#1}}

\let\oldaffiliation\affiliation
\def\affiliation#1{\oldaffiliation{\small #1}}
\usepackage{breqn}
\begin{document}

\preprint{APS/123-QED}

\title{OptPDE: Discovering Novel Integrable Systems via AI-Human Collaboration}

\author{Subhash Kantamneni}
\author{Ziming Liu}%

\author{Max Tegmark}
\affiliation{
  Department of Physics,
  Institute of Artificial Intelligence and Fundamental Interactions,
  Massachusetts Institute of Technology,
  Cambridge, USA
}

\date{\today}

\begin{abstract}

Integrable partial differential equation (PDE) systems are of great interest in  natural science, but are exceedingly rare and difficult to discover. To solve this, we introduce \textbf{OptPDE}, a first-of-its-kind machine learning approach that \textbf{Opt}imizes \textbf{PDE}s' coefficients to maximize their number of conserved quantities, $n_{\rm CQ}$, and thus discover new integrable systems. We discover four families of integrable PDEs, one of which was previously known, and three of which have at least one conserved quantity but are new to the literature to the best of our knowledge. We investigate more deeply the properties of one of these novel PDE families, $u_t = (u_x+a^2u_{xxx})^3$. 
Our paper offers a promising schema of AI-human collaboration for integrable system discovery: machine learning generates interpretable hypotheses for possible integrable systems, which human scientists can verify and analyze, to truly close the discovery loop.

\end{abstract}

\maketitle


\section{Introduction}


Integrable systems play a major role in physics 
and engineering because they are tractable, predictable and controllable~\cite{ hitchin2013integrable, isovervew}. Unfortunately, they are also extremely rare and difficult to discover~\cite{dunajski2012integrable,wiki:Integrable_system}. 
Traditional paper-pencil methods for discovering integrable systems focus on symbolic derivations and are highly ineffective, given the exponentially large search space of possible systems and conserved quantities (CQs), a defining property of integrable systems. The central research question of this paper is: \textit{can machine learning discover integrable systems?} To be specific, we focus on partial differential equations (PDEs). PDE integrability is more than an academic interest - for example,  optimizing the parameters of the PDE that describes plasma dynamics inside a tokamak~\cite{white2017theory} to increase its number of CQs ($n_{\rm CQ}$) can lead to greater controllability. We will not solve the million dollar question of nuclear fusion in this paper, but introduce a promising proof-of-concept in a simpler setup.

Our solution to integrable system discovery is \textit{OptPDE}, the first machine learning approach that optimizes PDE coefficients to maximize the number of conserved quantities, $n_{\rm CQ}$. To calculate $n_{\rm CQ}$ for any PDE, we introduce \textit{CQFinder}, which also retrieves symbolic formulas for these CQs given a symbolic basis specified by domain experts. There have been a number of works that use machine learning to discover conserved quantities from physical data \cite{aipoincare, hatrajectories, Mototake2021, PhysRevResearch.2.033499, modelagnostic, Lu2023, neuraldeflation, fluxfunction, learningdiscretize} and differential equations \cite{aipoincare2, SID}, but our method is the most interpretable for PDEs. More importantly, no existing method can actively optimize and design PDEs.
By maximizing $n_{\rm CQ}$ from CQFinder, OptPDE can discover integrable systems, as shown in Figure \ref{fig1}: OptPDE suppresses viscosity since viscosity harms conservation.

Armed with OptPDE, we carry out a massive search for integrable PDEs in the space of up to cubic polynomials of $u, u_x, u_{xx}$, and $u_{xxx}$, creating a general PDE with 33 trainable parameters. We independently run OptPDE with 5000 random sets of these parameters, and use dimensionality reduction to find clusters of solutions. 



We find that the solutions returned by OptPDE can be expressed as combinations of four families of PDEs, each with at least one conserved quantity. One of these families is $u_t = u_{xxx}$, which corresponds to (a special case of) the well known Korteweg–De Vries (KdV) equation, while three families are novel to the best of our knowledge. We investigate more deeply an exotic cluster of solutions corresponding to $u_t = (u_x+a^2u_{xxx})^3$.  We study the special case $a=0$ where $u_t=u_x^3$, which displays many interesting phenomena: wave-like solutions, existence of breaking (like for Burgers' equation), infinitely many CQs before breaking, power law decay of magnitudes after breaking and final convergence towards a triangular wave. 

While machine learning methods have been used to discover conserved quantities in the past \cite{LNNs, HNNs, DeLAN}, this paper presents the first example where AI and human scientists work in tandem to propose, verify, and interpret integrable systems, underscoring the promise of AI-human collaboration in physics and engineering.

\section{Method}

\subsection{Framework}

\begin{figure*}
  \includegraphics[width=\textwidth]{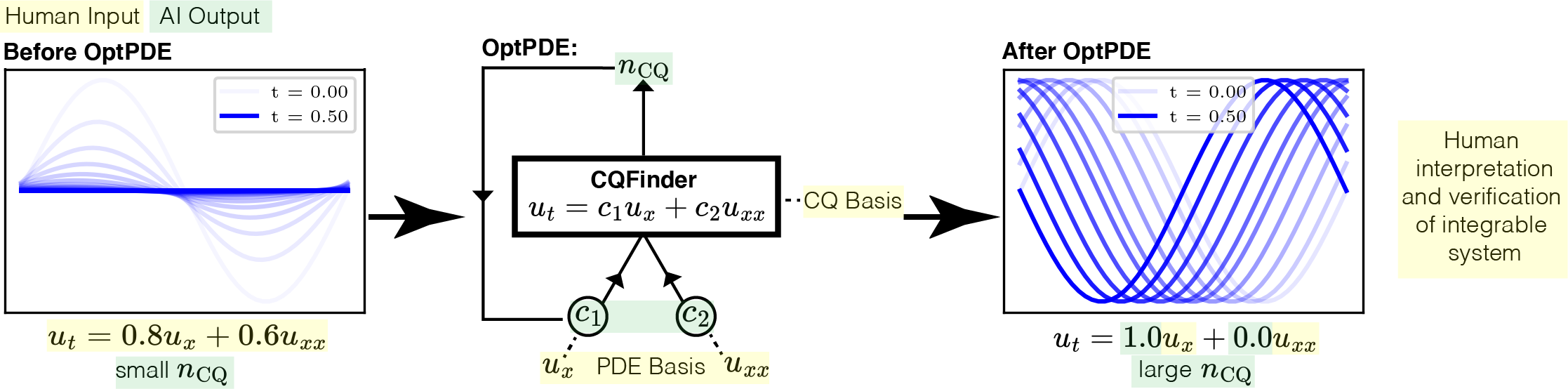}
  \caption{The OptPDE pipeline visualized. Given a basis of terms for a PDE, OptPDE optimizes the coefficients to maximize the number of conserved quantities (CQs) of the PDE, $n_{\rm CQ}$, the output of our CQFinder routine. Originally, $u$ decays and is not conserved, but OptPDE discovers coefficients that make $u$ more conserved by zeroing the diffusion term. The visualized example is trivial, but given an expansive PDE basis, OptPDE can help human scientists discover novel integrable systems. 
  We also highlight which steps of our pipeline are done by humans and AI, creating a collaborative workflow where humans propose hypotheses and interpret results, while AI carries out tedious computations.}
  \label{fig1}
\end{figure*}

We build our methodology in the following stages:
\begin{enumerate}
    \item \textbf{CQFinder} - Finds conserved quantities for a PDE.
    
    \textit{Input}: a) a PDE and b) a basis for CQs to be constructed from.
    
    \textit{Output}: $n_{\rm CQ}$, the number of CQs for the PDE and their symbolic forms.

    \item \textbf{OptPDE} - Uses $n_{\rm CQ}$ from CQFinder to discover integrable PDE systems.
    
    \textit{Input}: a) a basis for the PDE to be constructed from with initial coefficient values and b) a separate CQ basis.

    \textit{Output}: The coefficients for the PDE basis that maximize $n_{\rm CQ}$, the output of CQFinder. 
    
\end{enumerate}
We illustrate the pipeline in Figure \ref{fig1}. Note that our pipeline requires human scientists to work in tandem with AI by inputting a CQ and PDE basis, which require domain knowledge. 

\subsection{CQFinder: Discovering Conserved Quantities}

To construct OptPDE, we must first design CQFinder to accurately calculate the CQs of any PDE. To be specific, we consider a PDE that is first order in time with one spatial variable $x$, which has the form $u_t = f(u')$ where $u' = (u, u_x, u_{xx},u_{xxx},...)$ is a collection of $u$ and its spatial derivatives, and with the free boundary conditions $u(\pm\infty)=0$. 
We consider conserved quantities of the form $H = \int_{-\infty}^{\infty} h(u') dx$. For a quantity to be a CQ, it must be constant throughout the time evolution of $u$. 
We can express the time invariance of CQs as 

\begin{equation}
0 = \frac{dH}{dt} = \int_{-\infty}^{\infty} \frac{\partial h}{\partial t} dx = \int_{-\infty}^{\infty} \sum_{n} \frac{\partial h(u')}{\partial u_{nx} }\frac{\partial^n f(u')}{\partial x^n} dx,
\label{eq1}
\end{equation}
where $u_{nx} \equiv \frac{\partial^n u}{\partial x^n}$. Although this equation looks intimidating, we consider a simple setup where $h(u')$ is a linear combination of $K$ predefined basis functions, i.e., $h(u')=\sum_{i=1}^K \theta_ib_i(u')$. This parametrization makes Eq.~(\ref{eq1}) linear in the parameter $\boldsymbol \theta$. However, we need to handle two infinities here: (1) In theory, the linear equation must hold for any smooth $u$; in practice, we approximate this infinite set of functions by testing if Eq.~(\ref{eq1}) holds for a sampled subset $u_1, u_2, \cdots,u_P$. (2) In theory the integral is performed on $(-\infty, \infty)$; in practice, we approximate this with a finite range (imposing $u$ to be zero outside the range). After these two approximations, Eq.~(\ref{eq1}) is now a finite-dimensional linear system $\mat{G}\boldsymbol{\theta}=0$, with $\mat{G}\in\mathbb{R}^{P\times K}$ and $\boldsymbol{\theta}\in\mathbb{R}^K$. The derivation of $\mat{G}$ is left to Appendix \ref{deriveG}.

To obtain conserved quantities (which are solutions to $\mat{G}\boldsymbol{\theta}=0$), we apply singular value decomposition to \( \mathbf{G} = \mathbf{U\Sigma V}^T\), where \(\mathbf{U} \in \mathbb{R}^{P \times P} \) and \( \mathbf{V} \in \mathbb{R}^{K \times K} \) are orthogonal matrices and \( \mathbf{\Sigma} \in \mathbb{R}^{P \times K} \) is diagonal with singular values \( 0 \leq \sigma_1 \leq \sigma_2 \leq \cdots \). The number of conserved quantities is the number of vanishing singular values, $M$. Practically, we consider \( \sigma_i \) as effectively zero if $ \widetilde{\sigma_i} < \epsilon = 10^{-4} $, where $\boldsymbol{\widetilde{\sigma}}$ are normalized singular values such that $\sum_{i=1}^K \widetilde{\sigma_i}^2 = 1$. The corresponding solutions for these singular values are the first $M$ columns of $\mathbf{V}^T$, denoted $\mathbf{\Theta} = (\boldsymbol{\theta}^1, \boldsymbol{\theta}^2, \ldots, \boldsymbol{\theta}^M) \in \mathbb{R}^{K \times M}$. Note that $\mathbf{\Theta}$ forms a complete orthogonal basis of solutions. The $j^{\rm th}$ CQ of the system can be written as $h_j(u') = \sum_{i=1}^K \theta^j_i b_i(u')$.


We also create subprocesses in CQFinder for sparsification and identifying trivial solutions. We want sparse combinations of CQ bases as solutions since they are more interpretable to human scientists. 
We also want to identify trivial CQs (derivatives of some other function) and rule them out since they are conserved for any PDE. For example, if $uu_x$ is in our CQ basis, it will always be returned as a solution because $\int_{-\infty}^\infty uu_x dx = \int_{-\infty}^\infty d(u^2/2) = 0$ for all PDEs with free boundary conditions. Since they are non-essential for OptPDE, we defer their technical details to Appendix \ref{sparse} and \ref{trivial}.

\begin{figure*}[htbp]
  \includegraphics[width=\textwidth]{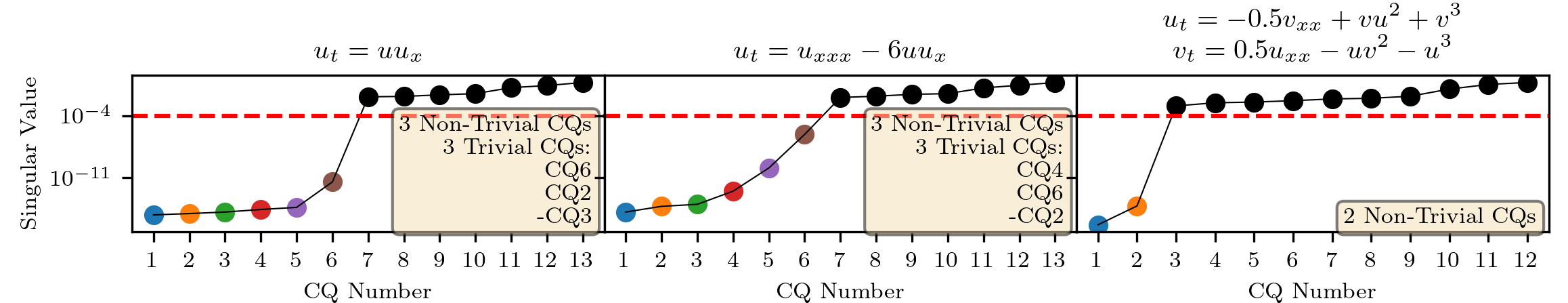}
  \caption{CQFinder results for Burger's equation (left), the Korteweg–De Vries equation (center) and the Non Linear Schr\"odinger equation (right) parameterized in real ($u$) and imaginary ($v$) parts. CQFinder correctly returns the number of CQs ($n_{\rm CQ}$) for each of these cases, in addition to their symbolic formulas (Appendix \ref{symbolic_forms}), and identifies trivial solutions that have a simple antiderivative. The ability for CQFinder to correctly identity $n_{\rm CQ}$ for PDEs lays the foundation for optimizing general PDE forms with OptPDE.}
  \label{singular_values}
\end{figure*}

\subsection{OptPDE: Discovering Integrable PDE Systems}

To be concrete, we parametrize the PDE as a linear combination of a pre-defined PDE basis, $u_t = \sum_i c_i f_i(u')$. CQFinder takes in a fixed PDE and outputs its number of conserved quantities $n_{\rm CQ}$. Since CQFinder is written in PyTorch \cite{pytorch}, it is in principle differentiable, i.e., we can identify which perturbations in the PDE's coefficients increase $n_{\rm CQ}$ via automatic differentiation. 
However, the biggest challenge for differentiability is that $n_{\rm CQ}$ is intrinsically discrete (e.g., a PDE can have 3 or 4 CQs, but not 3.7). To optimize the coefficients $c_i$ via backpropagation, our objective function ($n_{\rm CQ}$) must be differentiable. To solve this problem, we introduce a smoothed version of $n_{\rm CQ}$ using the sigmoid function:
\begin{equation}
    n_{\rm CQ} = \sum_{i=1}^K \frac{1}{1+\exp\left(\frac{\log(\sigma_i)-A}{B}\right)} , \quad \mathcal{L} = -n_{\rm CQ}
    \label{eq:L}
\end{equation}

where $\sigma_i$ is the $i^{\rm th}$ singular value from CQFinder and $A$ and $B$ are adjustable parameters. $A$ can be interpreted as the cutoff for a vanishing singular value, while $B$ defines how sharp the cutoff is. As $B$ goes to 0, the function becomes a step function, with value 1 for $\log \sigma_i < A$ and 0 for $\log \sigma_i > A$.  OptPDE uses backpropagation to optimize the coefficients $c_i$ to minimize $\mathcal{L}$, thus maximizing $n_{\rm CQ}$ and the integrabilty of the PDE.

\section{Results}


\subsection{Benchmarking CQFinder}

To verify that CQFinder works as claimed, we run it on three test systems: Burgers, Korteweg–De Vries (KdV), and Schr\"odinger's equations. In Figure~\ref{singular_values}, we show that the singular value curves display sharp phase transitions from small to large, making it clear cut to distinguish between vanishing and non-vanishing values. We show that CQFinder can not only correctly calculate the number of conserved quantities $n_{\rm CQ}$, but also obtain their symbolic formulas (Appendix \ref{symbolic_forms}). CQFinder's ability to accurately calculate $n_{\rm CQ}$ lays the foundation for OptPDE.

\subsection{AI: OptPDE Discovers Three Novel Integrable Systems}
\label{optpde_res}

\begin{figure}

  \includegraphics[width=1\linewidth]{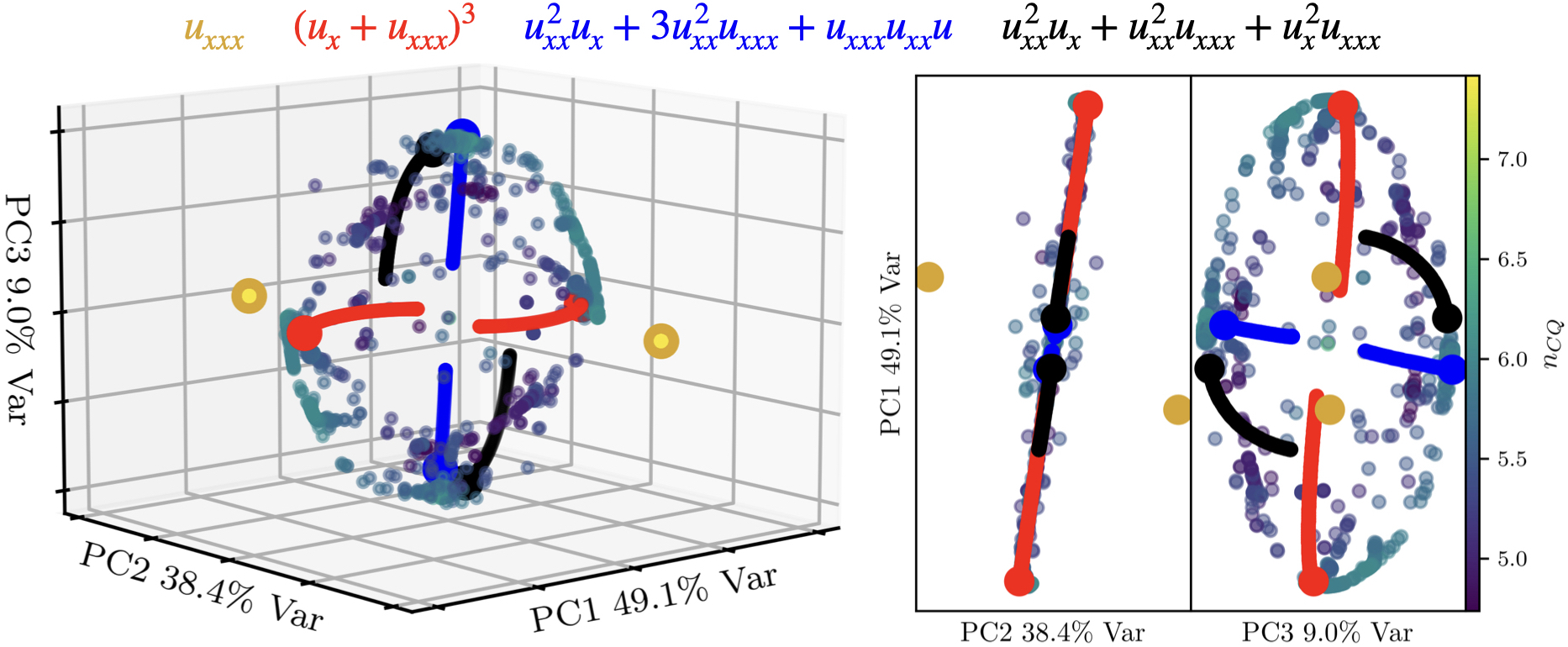}
  \caption{Each point on this 3D PCA corresponds to a PDE that OptPDE returns given PDE and CQ bases. We discover four families of PDEs with at least one conserved quantity, listed above. Note that these PDE families form bases for the subspace of OptPDE's results. For each family of PDEs, we use the transformation $x = ax'$ such that $u_{nx'} = a^nu_{nx}$ and plot the result in our PCA space for varying values of $a$. The $a = -1,1$ cases are shown with larger marker sizes.}
  \label{kdv_messy}
\end{figure}

\begin{figure*}
  \includegraphics[width=0.5\textwidth]{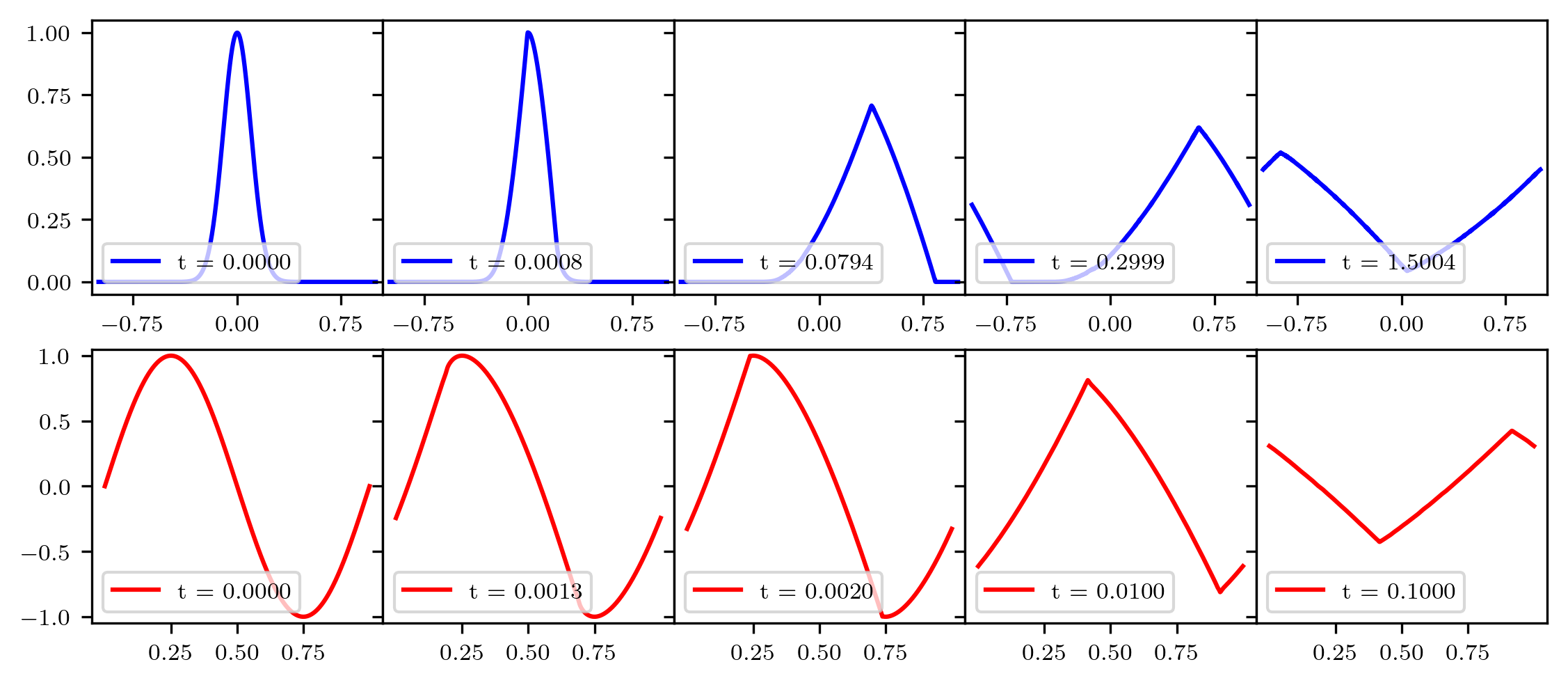}
  \includegraphics[width=0.4\textwidth]{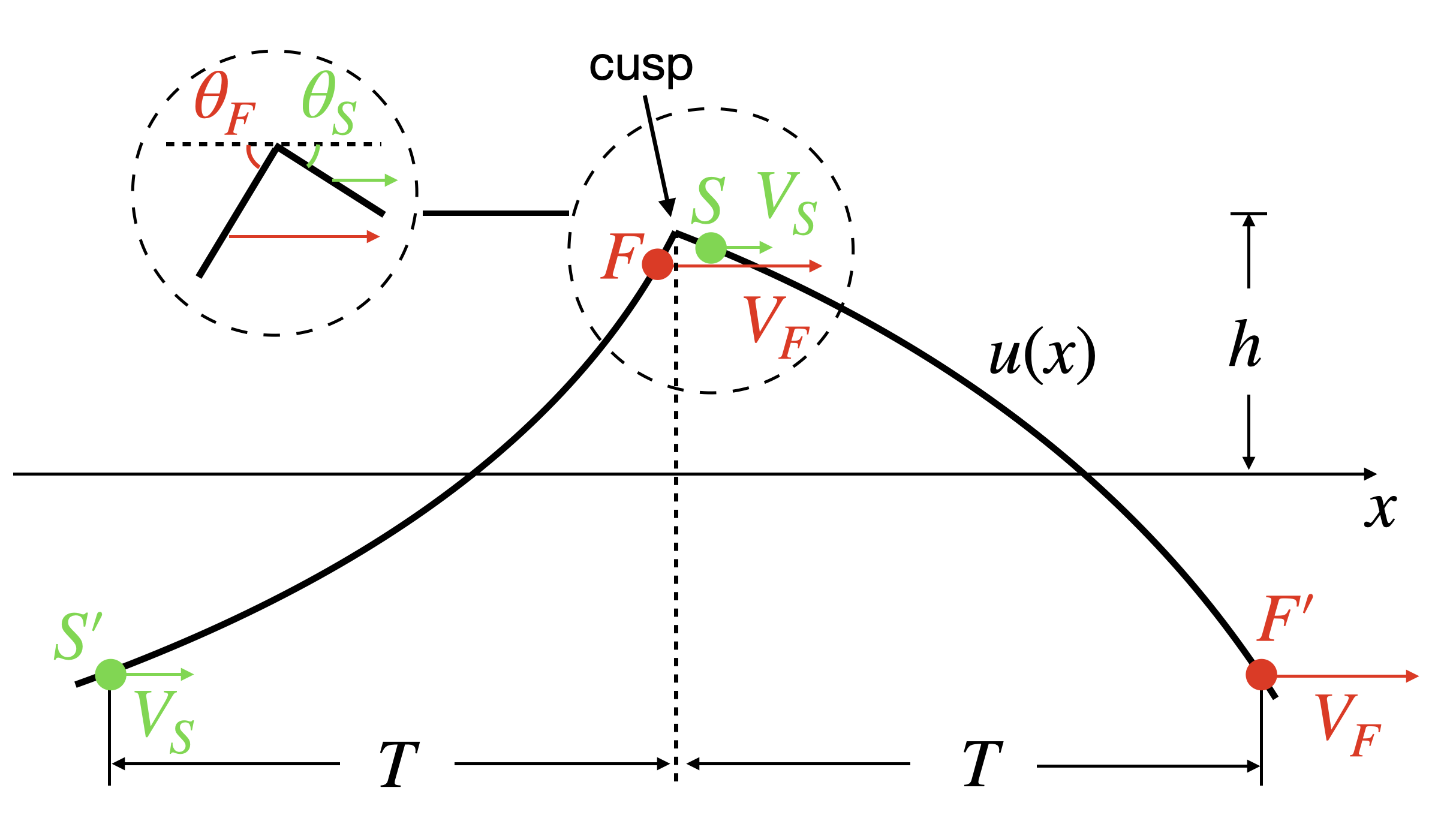}
  \caption{Human analysis of an AI-discovered integrable system. Left: We evolve a Gaussian (top) and sine wave (bottom) according to the equation $u_t = u_x^3$, which is a reduced form of the integrable PDE $u_t=(u_x+a^2u_{xxx})^3$ OptPDE discovered. There is a time at which a ``break" forms and both curves becomes non differentiable at a point. This break passes through the curves until they degrade into what visually appears to be straight lines. Right: A phenomenological model to explain the magnitude decay of the equation $u_t=u_x^3$ after breaking happens.}
  \label{pde_ev}
\end{figure*}

We can discover new integrable systems by using OptPDE to locate manifolds of integrable PDEs by maximizing $n_{\rm CQ}$. 
In Appendix \ref{warmup-example}, we detail a simple one parameter example of OptPDE: OptPDE learns to set a non-zero viscosity (diffusion) term to zero, since viscosity destroys conserved quantities.

For this paper, we choose our general PDE to be a single equation that can be written as $u_t = \sum_i c_ip(u')$, where $u' = u,u_x,u_{xx}, u_{xxx}$ and $p$ is a polynomial of degree up to 3. $u_t$ is composed of 34 terms, each with a trainable coefficient $c_i$. The CQ basis we use to evaluate $n_{\rm CQ}$ consists of up to third order polynomials of $u' = u,u_x,u_{xx}$ (CQ, PDE bases explicitly listed in Appendix \ref{bases_results}). To ensure that all coefficients $c_i$ do not go to 0 (since every function $h(u')$ is conserved for $u_t = 0$), we impose a normalization condition, such that $\sum c_i^2 = 1$. In practice, we use generalized spherical coordinates for our coefficients, which naturally imposes normalization (Appendix  \ref{sec:generalsphere}).

In OptPDE, we use $A = 0$, $B = 1000$, epochs = 25000, and a learning rate $10^{-3}$ with cosine annealing with $T_{max} = 5000$. Initially, all runs returned $u_t = u_x$, which represents the highly integrable advection equation, so we remove $u_x$ from the PDE basis to discover something more interesting. We run OptPDE with 5000 randomly chosen initialization locations for the remaining 33 parameters.

We analyze the results of OptPDE by visualizing the returned parameter values with a 3D PCA (explaining 96.5\% variance)~\cite{PCA}, shown in Figure \ref{kdv_messy}. The manifold structures of solutions are very interesting: two poles on either side and ring-like solutions in the middle. The two poles represent $u_t = u_{xxx}$, a reduced form of the integrable KdV equation, while the ring of solutions is more complex. We interpolate in this ring of solutions (details in Appendix \ref{interpret}) and find three families of PDEs that are a basis for the ring subspace, shown in Figure \ref{kdv_messy}. Each of these PDEs is, to the best of our knowledge, novel and inherently interesting due to their conserved quantities, which are shown in Appendix \ref{interpret}. We leave the study of two of these families to future work, and focus our analysis on the following PDE due to its compact form:
\begin{equation}
\resizebox{0.91\hsize}{!}{%
    $u_t = u_x^3+3au_x^2u_{xxx}+3a^2u_xu_{xxx}^2+a^3u_{xxx}^3 = \boxed{(u_x+a^2u_{xxx})^3}.$
}
    \label{discovery}
\end{equation}
We run CQFinder on the $a = 1$ case of this equation, and find that it has one non-trivial CQ: $h(u') = u_{x}^2-u_{xx}^2$. After a lengthy series of 
algebraic manipulations, we numerically and symbolically verify that $u_{x}^2-u_{xx}^2$ is indeed a CQ of $u_t = (u_x+u_{xxx})^3$ (see Appendix \ref{provecq}). Thus, OptPDE discovered a new PDE family which admits interesting conserved quantities: $u_t = (u_x+a^2u_{xxx})^3$.

\subsection{Human: Analyzing an AI-discovered System}
\label{interp}
It is now the responsibility of the human scientist to take the AI-discovered PDE family and interpret it. In this preliminary analysis, we restrict ourselves to analyzing the case $a\ll 1$, such that 
\begin{equation}
    u_t \approx u_x^3.
    \label{a=0}
\end{equation}

This special case represents a true integrable system and has an infinite number of CQs ($u_x^n$ is conserved for all $n$, shown in Appendix \ref{sec:infcq}). We plot the evolution of the PDE with both Gaussian and sinusoidal initial conditions in Mathematica, shown in Figure~\ref{pde_ev} left. Visually, the evolution appears to be a wave that degrades into linear components after a break time, when the wave becomes non-differentiable at a point. We derive a symbolic form of the break time and create a phenomological model for the post break time behavior of Eq.~(\ref{a=0}).


{\bf The Break Time} We use the method of characteristics to find when Eq.~(\ref{a=0}) ``breaks" and experiences a shock, visible in Figure \ref{pde_ev}. We first note that we can make Eq.~(\ref{a=0}) analogous to Burgers by differentiating both sides with respect to $x$. This results in $u_{xt} = 3u_x^2u_{xx}$, or 
$v_t = 3v^2v_x$ where $v \equiv u_x$. This new Burgers analogy represents a wave with speed $3v^2$. 
Employing the method of characteristics, with the characteristic equations $\frac{dx}{dt} = 3v^2$ and $\frac{du}{dt} = u_t + u_x\frac{dx}{dt} = u_t +3v^2u_x = 0$, we can trace out paths of constant $u$ and find the earliest time when two characteristics intersect. The mathematical details are left to Appendix  \ref{sec:breaktimemath}, but the resultant break time is 
$t_b = \mathrm{argmin}_{t_b > 0} \left\{ -\frac{1}{6u_xu_{xx}} \right\}$
which roughly matches up with our simulations in Appendix \ref{sec:breaktimemath}.

{\bf The Phenomenological Model} To understand the wave behavior after breaking, we aim to build a phenomenological model explaining the dynamics when the wave is close to a triangular wave. Assume that  the up part and down part can be regarded as linear plus a small quadratic correction $u_{xx}$ ($u_{xx}=C\ll 1$ on the left curve and $u_{xx}=-C$ on the right curve), shown in Figure~\ref{pde_ev} right. Denote the half period and height to be $T$ and $h$. $h$ decreases because at the cusp point, the left point $F$ travels faster to the right than the right point $S$ hence ``eats'' into the right curve.  Some derivations (in Appendix~\ref{app:phenom}) show that
$\frac{dh}{dt} \approx 24\frac{Ch^2}{T}.$
Further assuming that $C\propto h^\alpha$, we can derive that $h\propto t^{-\frac{1}{1+\alpha}}$. A special case is $\alpha=1$, when the curve is uniformly shrunk along height, we have $h\propto \frac{1}{\sqrt{t}}$, which agrees reasonably well with the sine wave case (shown in Appendix~\ref{sec:breaktimemath}).

{\bf Physical Understanding of Other Solutions}
From Figure \ref{kdv_messy}, we see that our solutions are high order and nonlinear, with cubic terms composed of third order derivatives. These can be daunting to apply physical intuition to, but we note that a third order derivative occurs in the KdV Equation, or if one derives the wave equation for a string with stiffness and another degree of resistance. Nonlinear PDEs are rare in physics but do occur - take the formula for air resistance at high speeds ($u_{tt} \propto u_t^2$). Thus, there is a basis for complex differential equations being useful in modeling physical phenomena, and we leave it to other scientists to make further strides in understanding these results.

\section{Conclusion}
We introduce an AI-human collaborative paradigm to simplify the discovery of integrable systems. We present OptPDE, an AI algorithm which takes in a human input basis of PDE terms and optimizes its coefficients to maximize the PDE's integrability. To construct OptPDE, we also introduce CQFinder, which automatically identifies CQs for a given PDE, from human input CQ basis. After running OptPDE with 5000 random initializations of PDE coefficient values, we discover four families of PDEs, one of which was previously known, and three of which are novel to the best of our knowledge.

We investigate one of these novel PDE families: $u_t = (u_x+a^2u_{xxx})^3$. We analyze the $a = 0$ case of this PDE, calculate its break time, and create a simple model to explain its power law decay in magnitude. We leave it open for future researchers to use this equation to model physical phenomena, to interpret the other novel PDE families we discovered, and to use OptPDE to discover more integrable PDEs. We hope that the collaborative paradigm between human scientists and AIs we introduced - where humans provide domain knowledge to the system, AI generates promising hypothesis, and humans perform interpretation and verification - inspires researchers to design similar AI schemas for other problems in physics.

\newpage

\bibliography{apssamp.bib}

\newpage

\appendix

\onecolumngrid


\section{Deriving the Matrix $\mat{G}$}
\label{deriveG}

We assume we're given a set of $K$ CQ basis terms $b_i(u') (i=1,2,\cdots,K)$, and want to discover which linear combinations of CQ basis terms are conserved for the equation $u_t = f(u')$, i.e., $h(u') = \sum_{i=1}^K \theta_i b_i(u')$. We can rewrite the conservation condition, Eq.~(\ref{eq1})  as 
\begin{equation}
0 = \sum_{i=1}^K \theta_i \int  \pdv{b_i}{u}f + \pdv{b_i}{u_x}\pdv{f}{x} + ... \pdv{b_i}{u_{nx}}\pdv[n]{f}{x}dx.
\label{cons_condition}
\end{equation}
By interpreting the integral in Eq.~(\ref{cons_condition}) as a summation over $N_p$ points in $x$-space, we can rewrite Eq.~(\ref{cons_condition}) as 
\begin{equation}
0 = \sum_{i=1}^K \theta_i \sum_{N_p \textrm{ x-points}} g(u'(x)),
\label{eq2}
\end{equation}
where $g(u') = \pdv{b_i}{u}f + \pdv{b_i}{u_x}\pdv{f}{x} + ... \pdv{b_i}{u_{nx}}\pdv[n]{f}{x}$ and we drop the constant factor of $dx$. In our setup, $b(u')$ and $f(u')$ are known, so $g(u')$ is also known. To find $\theta$, we note that Eq.~(\ref{eq2}) should hold for any $u'$. 

To leverage this, we generate $P$ curves of $u$ with a Gaussian mixture model of randomized parameters. Our Gaussian mixture model has the form $u_p = \sum_{g=1}^{N_g} A_g\mathrm{exp}(-\frac{(x-\mu_g)^2}{2\sigma_g^2})$. We use $N_g = 10$, $\mu = U[-3,3], \sigma = 1.5, A = U[-5,5]$ and generate $N_p = 1000$ points in $x = U[-10,10]$. We generate the rest of $u_p'$ by taking spatial derivatives of $u_p$ symbolically with the Python library SymPy \cite{sympy}. $\mu$ and $\sigma$ were chosen such that all derivatives in $u_p'$ are 0 at the interval ends for all $p$ curves. We can then write

\begin{equation}
\underbrace{\left(
\begin{array}{cccc}
\sum g_1(u'_1) & \sum g_2(u'_1) & \cdots & \sum g_K(u'_1) \\
\sum g_1(u'_2) & \sum g_2(u'_2) & \cdots & \sum g_K(u'_2) \\
\vdots & \vdots & \ddots & \vdots \\
\sum g_1(u'_P) & \sum g_2(u'_P) & \cdots & \sum g_K(u'_P) \\
\end{array}
\right)}_{\mathbf{G}}
\underbrace{\left(
\begin{array}{c}
\theta_1 \\
\theta_2 \\
\vdots \\
\theta_K \\
\end{array}
\right)}_{\boldsymbol{\theta}}
= 0.
\label{CQFinder}
\end{equation}

\section{CQFinder Test Examples}
\label{symbolic_forms}

\begin{figure*}[htbp]
  \includegraphics[width=\textwidth]{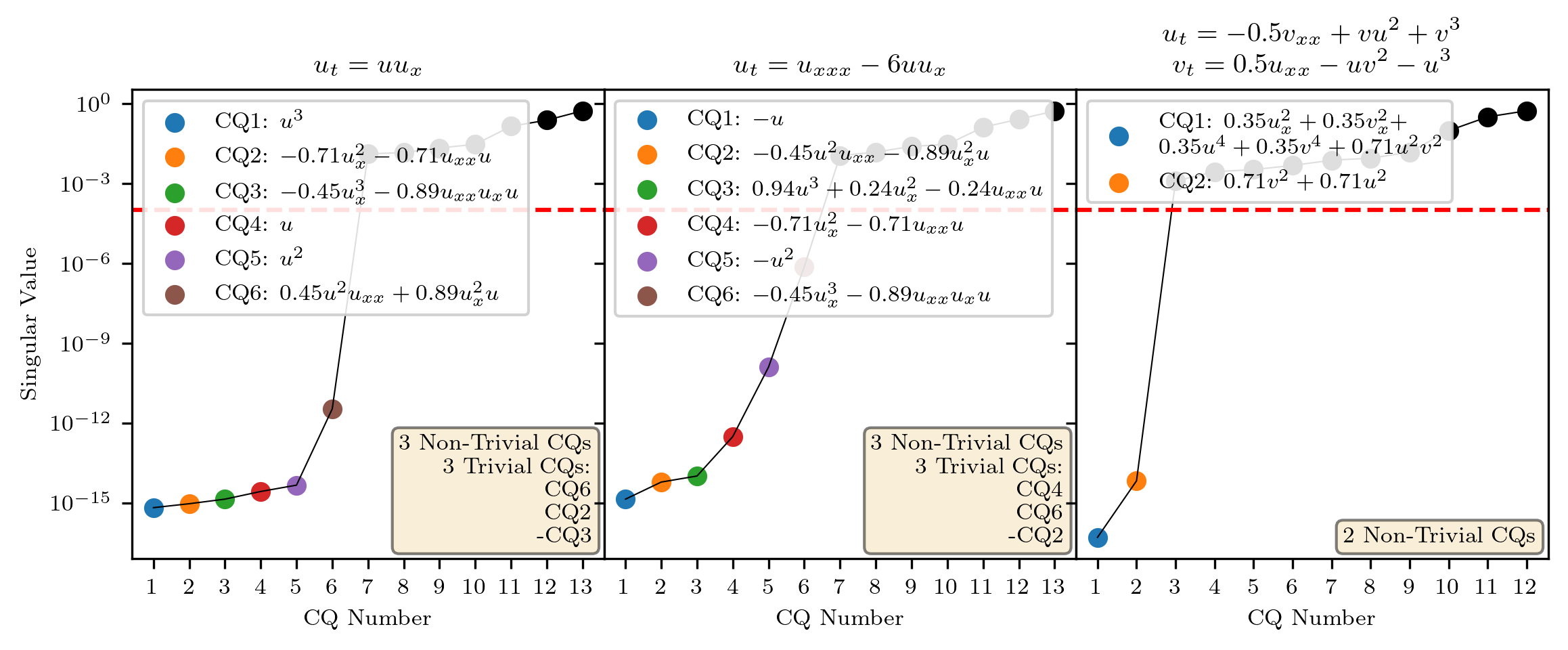}
  \caption{CQFinder results for Burger's equation (left), the Korteweg–De Vries equation (center) and the Non Linear Schr\"odinger equation (right) parameterized in real ($u$) and imaginary ($v$) parts. Given a user-inputted CQ basis, CQFinder uses linear regression to determine the conserved quantities of a PDE  using a singular value threshold of $10^{-4}$. CQFinder also computes which of these conserved quantities are a trivial antiderivative of a simple function. We rediscover all expected conserved quantities for these examples. Note that the human scientist is still required for the interpretation of these results. For example, CQ3 for KdV is listed as $0.94u^3+0.24u_x^2-0.24u_{xx}u$. The human scientist must realize that you can add any function with a symbolic antiderivative to a CQ, note that $u_x^2+uu_x = \frac{d}{dx}uu_x$, so CQ2 is $0.94u^3+0.48u_x^2$. This closely matches KdV's third integral of motion, $2u^3+u_x^2$, but it is once again up to the human scientist to look past the small difference in coefficients and identify them as the same.}
  \label{singular_values_appendix}
\end{figure*}

{\bf{The Burgers equation}} is given by $u_t = uu_x$ \cite{Burgers1948}. It can be shown that all $h(u') = u^n$ for $n>1$ are conserved quantities of Burgers equation. From Eq.~(\ref{eq1}), we see that $\frac{dH}{dt} = \int (\partial_u u^n)uu_x = u^{n+1}$ which evaluates to 0 using the zero boundary conditions. We rediscover all inputted $u^n$ in Figure \ref{singular_values_appendix}.

{\bf{The KdV equation}} is $u_t = u_{xxx}-6uu_x$ and models waves on shallow water surfaces \cite{Korteweg1895}. The KdV equation has an infinite number of integrals of motion \cite{Miura1968}. The first three integrals of motion are $u, u^2, 2u^3-u_x^2$, which we rediscover in Figure \ref{singular_values_appendix}. 

{{\bf The Non Linear Schr\"odinger equation}} (NLSE) is $i \psi_t + \frac{1}{2} \psi_{xx} - |\psi|^2 \psi = 0$ and conserves the norm squared ($ M = \int | \psi |^2 \, dx $) and the Hamiltonian ($ H = \int \left( \frac{1}{2} |\psi_x|^2 + \frac{1}{2} |\psi|^4 \right) \, dx $) \cite{barrett2013local, gutkin1985conservation}. We extend our formulation to systems of PDEs (see Appendix \ref{sec:systems}) and break $\psi$ into real and imaginary parts. Using $\psi = u+iv$, the NLSE becomes $u_t = -\frac{1}{2}v_{xx}+v(u^2+v^2)$ and $v_t = \frac{1}{2}u_{xx}-u(u^2+v^2)$. The conserved quantities, $M$ and $H$, can be written as $M =  \int (u^2 + v^2) \, dx$ and $H = \int \frac{1}{2} (u_x^2 + v_x^2) + \frac{1}{2} (u^2 + v^2)^2 \, dx$. We rediscover these in Figure \ref{singular_values_appendix}.

\section{Extending CQFinder to Systems of PDEs}
\label{sec:systems}
Many notable equations in physics, like the Schr\"odinger equation, are represented as systems of PDEs. We want to extend CQFinder to systems of PDEs to make it more generally applicable. Let's assume we're given a system of $n$ equations in the form
\begin{equation}
\begin{split}
u^1_t &= f({u^1}', \ldots, {u^n}') \\
\vdots \\
u^n_t &= f^n({u^1}', \ldots, {u^n}'),
\label{system}
\end{split}
\end{equation}
where ${u^i}' = (u^i,u^i_x,\ldots,u^i_{xxx\ldots})$. 
We can express our conserved quantities in a similar fashion to the single PDE case.
\begin{equation}
H = \int h({u^1}', {u^2}', \ldots, {u^n}') dx \approx \sum_{N_p} h({u^1}', {u^2}', \ldots, {u^n}'),
\label{CQ_def}
\end{equation}
where we define our integral as the sum over $N_p$ evenly spaced $x$-points. Then, combining Eq.\eqref{system} and Eq. \eqref{CQ_def} and forcing $\frac{dH}{dt} = 0$ to be conserved, we have

\begin{equation}
    \begin{split}
        \frac{dH}{dt} &= 0 = \sum_{N_p}\left(\frac{dh}{du^1}\frac{du^1}{dt} + \ldots + \frac{dh}{du^1_{xxx\ldots}}\frac{du^1_{xxx\ldots}}{dt}\right) \\
        &+ \ldots + \left(\frac{dh}{du^n}\frac{du^n}{dt} + \ldots + \frac{dh}{du^n_{xxx\ldots}}\frac{du^n_{xxx\ldots}}{dt}\right).
    \end{split}
\end{equation}
We can then write this as 
\begin{equation}
    \sum_{N_p}\sum^{n}_{i=1} \frac{dh}{du^i}f^i+\ldots+\frac{dh}{du^i_{xxx\ldots}}f^i_{xxx\ldots} = 0,
    \label{doublesum}
\end{equation}
where we leverage that $\frac{du^i}{dt} = f^i$. Now, we can introduce our $K$ multi-variable CQ basis functions and posit that  
\begin{equation}
h = \sum_{j=1}^K \theta_{j} b_j({u^1}', \ldots, {u^n}').
\label{CQ_basis}
\end{equation}
We can use Eq. \eqref{CQ_basis} to rearrange and rewrite Eq. \eqref{doublesum} as
\begin{equation}
\begin{split}
    \sum_{j=1}^K \theta_j \sum_{N_p} \sum_{i=1}^n \frac{db_j}{du^i}f^i+\ldots+\frac{db_j}{du^i_{xxx\ldots}}f^i_{xxx\ldots} = 0.
\end{split}
\end{equation}

Analogous to the single PDE case, we can define $g_j^i = \frac{db_j}{du^i}f^i+\ldots+\frac{db_j}{du^i_{xxx\ldots}}f^i_{xxx\ldots}$. Evaluating this with $P$ random constructions of $u^i$ and defining $x_s = ({u_s^1}',{u_s^2}',\ldots,{u_s^n}')$, we can write the conservation condition as

\begin{equation}
    \left( \begin{array}{c}
    \Sigma_{N_p} \Sigma_i^n g_1^i(x_1) \cdots \Sigma_{N_p} \Sigma_i^n g_K^i(x_1) \\
    \Sigma_{N_p} \Sigma_i^n g_1^i(x_2) \cdots \Sigma_{N_p} \Sigma_i^n g_K^i(x_2) \\
    \vdots \\
    \Sigma_{N_p} \Sigma_i^n g_1^i(x_P) \cdots \Sigma_{N_p} \Sigma_i^n g_K^i(x_P)
\end{array} \right)
\left( \begin{array}{c}
    \theta_1 \\
    \theta_2 \\
    \vdots \\
    \theta_K
\end{array} \right) = 0.
\end{equation}
This reduces to

\begin{equation}
\sum_{i=1}^n \underbrace{\left( 
\begin{array}{ccc}
    \sum_{N_p} g_1^i(x_1) & \cdots & \sum_{N_p} g_K^i(x_1) \\
    \sum_{N_p} g_1^i(x_2) & \cdots & \sum_{N_p} g_K^i(x_2) \\
    \vdots & \ddots & \vdots \\
    \sum_{N_p} g_1^i(x_P) & \cdots & \sum_{N_p} g_K^i(x_P)
\end{array} 
\right)}_{\boldsymbol{G}_i}
\underbrace{\left( 
\begin{array}{c}
    \theta_1 \\
    \theta_2 \\
    \vdots \\
    \theta_K
\end{array} 
\right)}_{\boldsymbol{\theta}}
= 0.
\end{equation}

Thus, procedurally we can simply calculate the matrix $\bm{G}_i$ for each equation in our system $u^i_t = f({u^1}', \ldots, {u^n}')$ and then summate them to get our final matrix $\bm{G}$. We can then use the CQFinder algorithm as normal to discover conserved quantities for systems of PDEs.

\section{Sparsifying Solutions for Interpretability in CQFinder}
\label{sparse}
We would like our CQFinder solutions $\mathbf{\Theta}$ to be as sparse as possible in order to maximize their interpretability. Note that if $\mathbf{R} \in \mathbb{R}^{M \times M}$ is an orthogonal matrix, $\mathbf{\Theta}' = \mathbf{R}\mathbf{\Theta}$ is a set of valid solutions to Eq. \eqref{CQFinder}. We want to choose $\mathbf{R}$ such that most values in $\mathbf{\Theta}'$ are 0. Thus, we use

\begin{equation}
\mathbf{R}^* = \underset{\mathbf{R}^T\mathbf{R}=\mathbf{I}}{ \arg \min} \ \| \mathbf{\Theta} \mathbf{R} \|_1, \quad \mathbf{\Theta}' = \mathbf{\Theta} \mathbf{R}^*, \tag{5}
\end{equation}
where $\| \mathbf{M} \|_1$ represents the $L_1$-norm of the matrix $\mathbf{M}$. In practice, we use the SciPy's Limited-memory BFGS optimization method to calculate $\mathbf{R}^*$ \cite{Liu1989, 2020SciPy-NMeth}.

\section{Identifying Trivial Solutions in CQFinder}
\label{trivial}
Because we impose zero boundary conditions for all $u_p'$ in CQFinder, any $h(u_p')$ that has an anti-derivative is conserved, with value 0:
\begin{equation}
    \int_i^f h(u_p')dx = H(u'_{pf})-H(u'_{pi}) = 0,
\end{equation}
where $H$ is the anti-derivative of $h$ and $u'_{pf}$ and $u'_{pi}$ are the boundary values of $u_p'$, which we've constructed to be 0.

These solutions are uninteresting in that they are conserved for any and every PDE. To address this, we identify trivial CQs from our solution set with a procedure similar to Eq. \eqref{CQFinder}. If a CQ is trivial, then $\int h(u_p') dx \approx \sum_x h(u_p') = 0$ for all $u_p'$, so we can write

\begin{equation}
\underbrace{\left(
\begin{array}{cccc}
\sum h_1(u'_1) & \sum h_2(u'_1) & \cdots & \sum h_M(u'_1) \\
\sum h_1(u'_2) & \sum h_2(u'_2) & \cdots & \sum h_M(u'_2) \\
\vdots & \vdots & \ddots & \vdots \\
\sum h_1(u'_P) & \sum h_2(u'_P) & \cdots & \sum h_M(u'_P) \\
\end{array}
\right)}_{\mathbf{T}}
\underbrace{\left(
\begin{array}{c}
\theta_1 \\
\theta_2 \\
\vdots \\
\theta_k \\
\end{array}
\right)}_{\boldsymbol{\theta}_T}
= 0.
\label{eq3}
\end{equation}

Trivial CQs are formed in the null space of $\mathbf{T}$ and can be identified and sparsified similarly to Eq. \eqref{CQFinder}. If the rank of the null space of $\mathbf{T}$ is $M_T$, then the number of non-trivial CQs is $M-M_T$.

\section{Bases Used for Results}
\label{bases_results}
For the CQFinder results we present on Burgers, KdV, and Schr\"odingers equation, we use the following CQ bases.
\begin{table}[h]
\begin{tabular}{cc}
\multicolumn{1}{l}{\textbf{Burgers/KdV}} & \multicolumn{1}{l}{\textbf{NLSE}} \\
$u$                                       & $u$                               \\
$u^2$                                     & $u^3$                             \\
$u^2u_{xx}$                               & $u_{x}^3$                         \\
$u^3$                                     & $u_{xx}^2$                        \\
$u_{x}^2$                                 & $u_{xx}^3$                        \\
$u_{x}^2u$                                & $v^2$                             \\
$u_{x}^3$                                 & $u_{x}^2$                         \\
$u_{xx}^2$                                & $v_{x}^2$                         \\
$u_{xx}^2u$                               & $u^4$                             \\
$u_{xx}^2u_{x}$                           & $v^4$                             \\
$u_{xx}^3$                                & $u^2v^2$                          \\
$u_{xx}u$                                 & $u^2$                             \\
$u_{xx}u_{x}u$                            & \multicolumn{1}{l}{}              \\
\multicolumn{1}{l}{}                      & \multicolumn{1}{l}{}             
\end{tabular}
\end{table}

For the OptPDE run results we present in Section \ref{optpde_res}, we use the Burgers/KdV CQ basis in the above table, and our PDE basis is up to third order polynomials of $u' = (u,u_x,u_{xx}, u_{xxx})$. Explicitly, our PDE basis is 
\begin{align*}
    u, & u^2, u^2u_{x}, u^2u_{xx}, u^2u_{xxx}, u^3, u_{x}, u_{x}^2, u_{x}^2u, \\
    u_{x}^2 & u_{xx}, u_{x}^2u_{xxx}, u_{x}^3, u_{x}u, u_{xx}, u_{xx}^2, u_{xx}^2u, \\
    u_{xx}^2 & u_{x}, u_{xx}^2u_{xxx}, u_{xx}^3, u_{xx}u, u_{xx}u_{x}, u_{xx}u_{x}u, \\
    u_{xxx}, & u_{xxx}^2, u_{xxx}^2u, u_{xxx}^2u_{x}, u_{xxx}^2u_{xx}, u_{xxx}^3, \\
    u_{xxx}u, & u_{xxx}u_{x}, u_{xxx}u_{x}u, u_{xxx}u_{xx}, u_{xxx}u_{xx}u, u_{xxx}u_{xx}u_{x}.
\end{align*}

This PDE basis contains 34 terms. When running OptPDE on this basis, the only solutions returned are $u_t = u_x$, which is the already well-studied advection equation. To retrieve more interesting, novel systems, we remove $u_x$ from our basis and run OptPDE with 33 terms in the PDE basis. 

\section{A Warm Up Example for OptPDE: KdV with Diffusion}
\label{warmup-example}

We are given the KdV equation with a small diffusion term with a variable coefficient $k$: $u_t = u_{xxx}-6uu_x+ku_{xx}$. Can we find the value of $k$ that maximizes the number of conserved quantities of the system, $n_{\rm CQ}$? 

\begin{figure}
  \includegraphics[width=0.5\columnwidth]{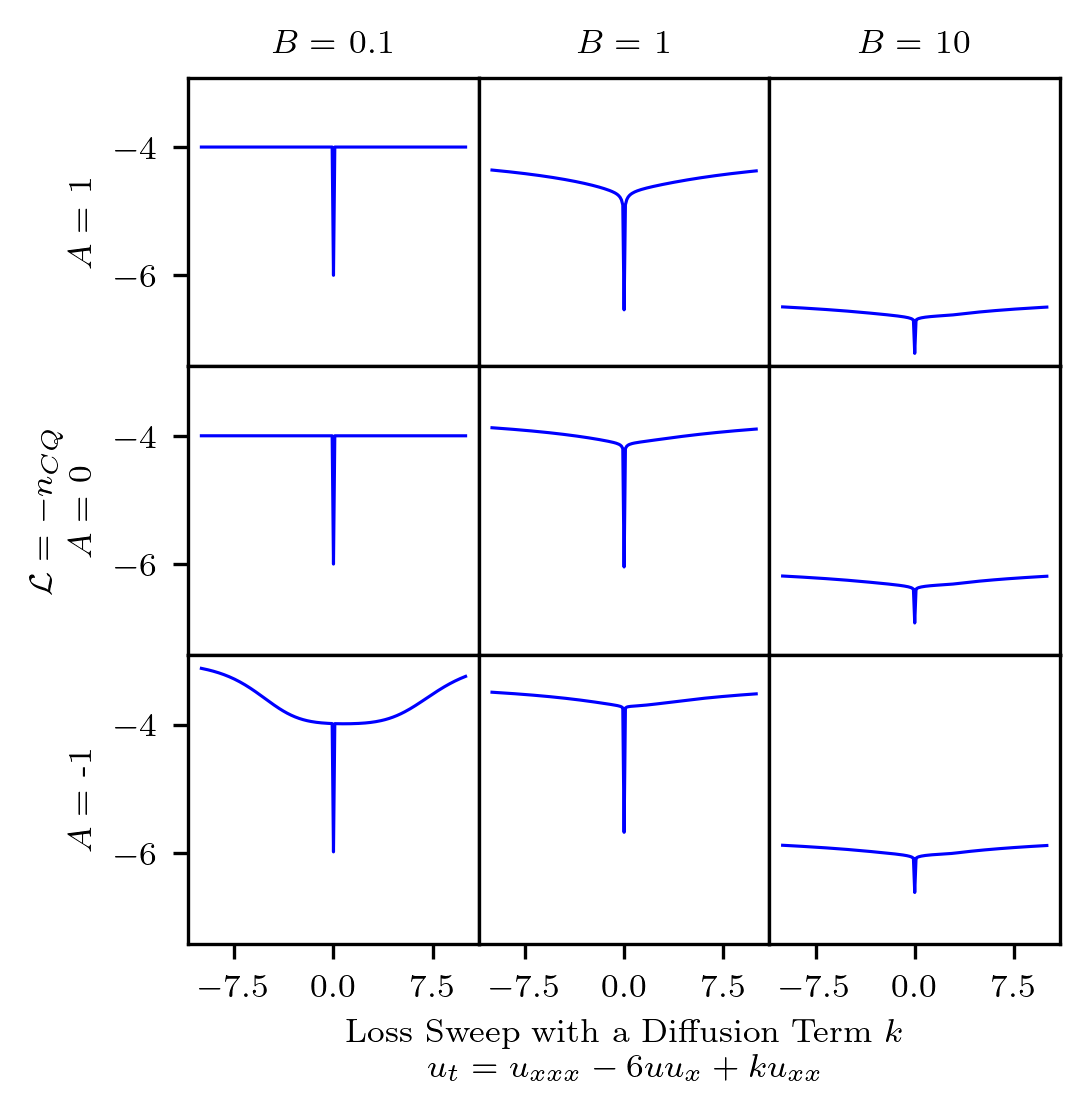}
  \caption{For values in $k$ in $[-10,10]$ we calculate $\mathcal{L} = -n_{\rm CQ} = -\sum_{i=1}^K {(1+e^{\frac{\log{\sigma_i}-A}{B}})}^{-1}$ for the equation $u_t = u_{xxx} - 6uu_x +ku_{xx}$ with a variety of values for $A, B$. When $k = 0$, there is no diffusion and the equation is simply the KdV equation, which has the most conserved quantities. This loss landscape should have two properties: 1) it must be smooth and differentiable and 2) it should accurately reflect the number of CQs in the system. When $A$ is smaller, our threshold becomes ``stricter" and more accurate, and when $B$ increases the range of loss values we receive becomes smaller.}
  \label{sweep}
\end{figure}

Notably, the conserved quantities for the KdV equation listed in Appendix \ref{symbolic_forms} only hold for a zero diffusion term, so $n_{\rm CQ}$ is maximized when $k = 0$. First, we manually sweep through $k$ from [-10,10] in steps of 0.1. As we expect, $\mathcal{L}$ from Eq. \eqref{eq:L} is maximized when $k = 0$, and is plotted in Figure \ref{sweep} for a variety of $A, B$ values.

We can also use OptPDE to minimize $\mathcal{L}$ from Eq. \ref{eq:L} with backpropagation instead of a manual sweep over parameter space. OptPDE has the advantage of being scalable - parameter sweeps become infeasible when there are many parameters and thus an exponential search space, but this is the type of environment that machine learning techniques like OptPDE flourish in. 

\begin{figure}
  \includegraphics[width=0.5\columnwidth]{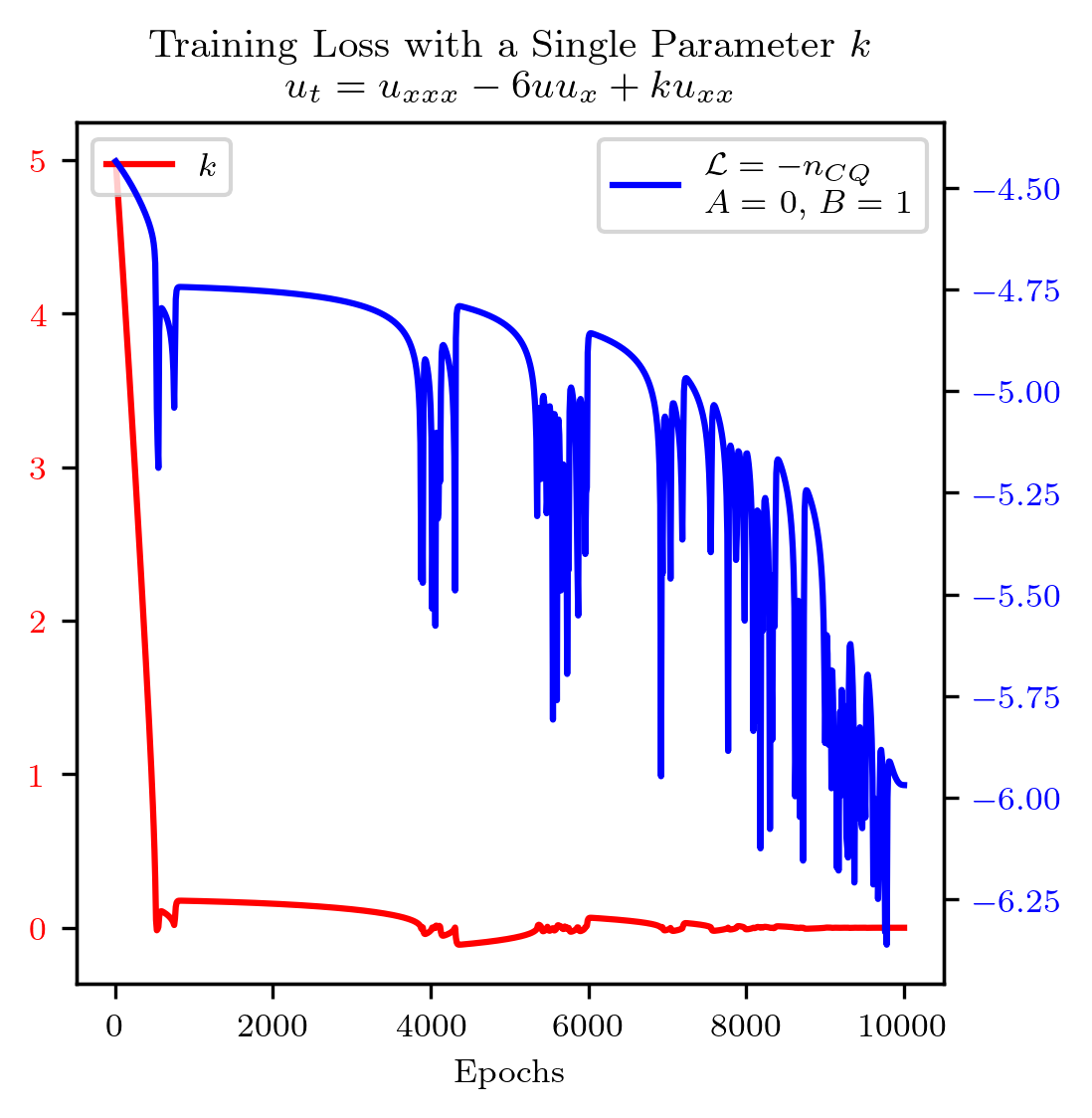}
  \caption{We run our model on the KdV equation with a single trainable parameter, $k$, with $\mathrm{epochs} = 10000$, learning rate = $5e-3$ with cosine annealing and $A = 0, B = 1$ in the loss function. As expected, the lowest loss occurs when $k = 0$, and the equation reduces to the KdV equation. Notably, the model returns a lowest loss near $\mathcal{L} = -n_{\rm CQ} = -6$, which is expected because the KdV equation has 6 conserved quantities in the given CQ basis (three trivial, three non-trivial).}
  \label{single_loss_curve}
\end{figure}

We use $A = 0$ and $B = 1$ in OptPDE's loss function to create a smooth, easily differentiable loss landscape with roughly accurate $n_{\rm CQ}$ values. The parameter $k$ quickly goes to 0, which is shown in Figure \ref{single_loss_curve} along with the loss curve for this run. We use $\mathrm{epochs} = 10000$, learning rate = $5e-3$, and cosine anneal the learning rate.

Note that when we run OptPDE with a large set of PDE bases, we use $A = 0, B = 1000$ to carry out the optimization. The $n_{\rm CQ}$ values returned are less accurate, but the overall loss landscape is smoother for larger $B$ as shown in Figure \ref{sweep}, which is important for a high dimensional optimization problem.

\section{Generalized Spherical Coordinates for OptPDE Coefficient Normalization}
\label{sec:generalsphere}

We want the coefficients in our OptPDE optimization loop to be normalized to 1, otherwise they will all go to 0, since every function is conserved for $u_t = 0$. Spherical coordinates in three dimensions are naturally normalized, and we use the extension of this definition to $n$ dimensions to impose normalization on our problem~\cite{wiki:Spherical_coordinate_system}.

For our $n$ dimensional space, we have one radial coordinate $r$ and $n-1$ angular coordinates $\phi_1, \phi_2, \ldots, \phi_{n-1}$, where $\phi_i$ for $1\leq i\leq n-2$ ranges from $[0,\pi]$, and $\phi_{n-1}$ ranges from $[0, 2 \pi]$. For normalization, we impose $r = 1$.

So, if our PDE could be expressed as 
$u_t = \sum_{i=1}^n c_ip_i(u')$, where $c_i$ are Cartesian coordinates, upon initialization of OptPDE we create random, unbounded values of $c_i$ and convert them to spherical coordinates using 

\begin{equation}
\begin{aligned}
r &= \sqrt{x_n^2 + x_{n-1}^2 + \ldots + x_2^2 + x_1^2}, \\
\phi_1 &= \atan2(\sqrt{x_n^2 + x_{n-1}^2 + \ldots + x_2^2}, x_1), \\
\phi_2 &= \atan2(\sqrt{x_n^2 + x_{n-1}^2 + \ldots + x_3^2}, x_2), \\
&\vdots \\
\phi_{n-2} &= \atan2(\sqrt{x_n^2 + x_{n-1}^2}, x_{n-2}), \\
\phi_{n-1} &= \atan2(x_n, x_{n-1}),
\end{aligned}
\end{equation}
where atan2 is the two-argument arctangent function.

We divide out by $r$ to ensure our coordinates are normalized. Holding $r = 1$, we then optimize the $n-1$ angular coordinates throughout OptPDE. At every step of OptPDE with a certain set of angular coordinates, we use the inverse transformation to run CQFinder with normal Cartesian coordinates:

\begin{equation}
\begin{aligned}
x_1 &= \cos(\phi_1), \\
x_2 &= \sin(\phi_1) \cos(\phi_2), \\
x_3 &= \sin(\phi_1) \sin(\phi_2) \cos(\phi_3), \\
&\vdots \\
x_{n-1} &= \sin(\phi_1) \cdot \ldots \cdot \sin(\phi_{n-2}) \cos(\phi_{n-1}), \\
x_n &= \sin(\phi_1) \cdot \ldots \cdot \sin(\phi_{n-2}) \sin(\phi_{n-1}),
\end{aligned}
\end{equation}
and use the results of CQFinder to optimize the angular coordinates $\phi_i$ in OptPDE.

\section{Interpreting the Ring of Solutions in Figure \ref{kdv_messy}}
\label{interpret}

The two poles in Figure \ref{kdv_messy} correspond to the solution $u_t = u_{xxx}$, which popped out as a result from OptPDE without further manipulation required.

In theory, every point in the ring of solutions in Figure \ref{kdv_messy} should correspond to a PDE with CQs. In practice, points in the ring are linear combinations of underlying PDE families with CQs, while not having CQs themselves. The key challenge is thus: how can we discover the underlying PDE families with conserved quantities?

In practice, we adopted the following approach:
\begin{enumerate}
    \item For all solutions OptPDE finds, threshold PDE bases coefficients of $< 0.1$ to $0$
    \item After doing Step 1 on all 5000 OptPDE solutions, roughly 20 families of solutions crystallize
    \item Look individually at each family ranked by frequency. Via the eye test, tease out the coefficients from the floating point approximations. An inductive bias that seems to hold given our solutions is that coefficents of integrable PDEs are often rational.
    \item Test out the suspected form of the PDE with CQFinder. If a non-trivial CQ is returned, then our PDE is integrable. If not, vary coefficient values until it's integrable. One helpful technique was thresholding increasingly larger coefficient values to 0.
\end{enumerate}

\begin{figure}
    \centering
    \includegraphics[width = \textwidth]{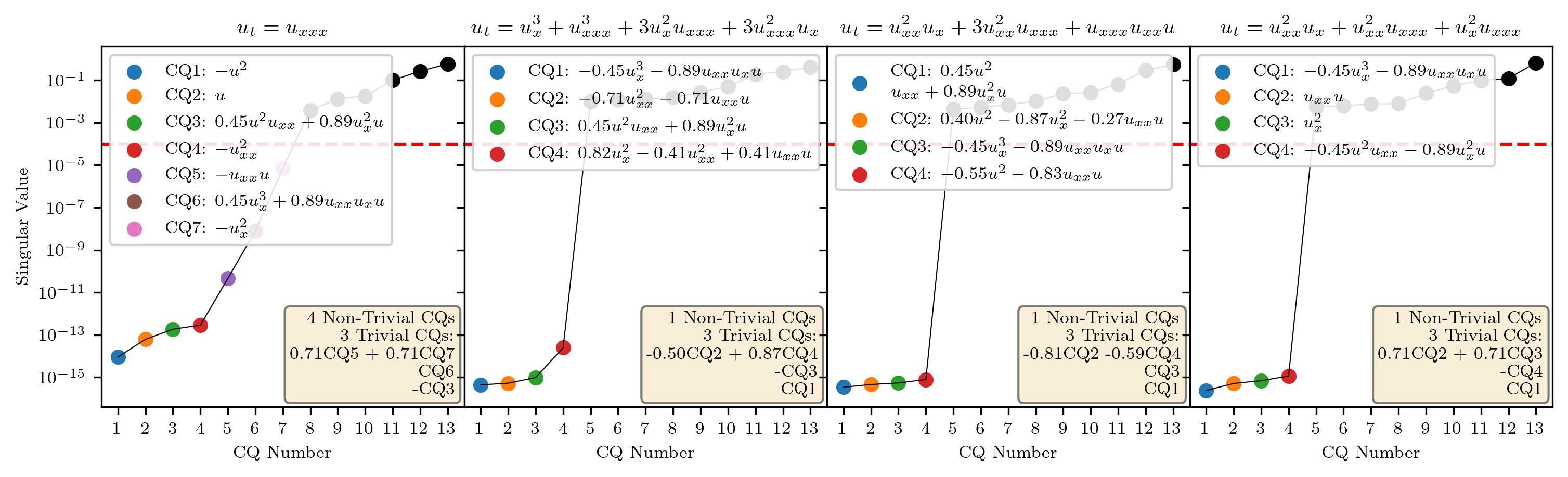}
    \caption{CQs for the four families of PDEs discovered by OptPDE (shown in Figure \ref{kdv_messy}). Each family has at least one CQ. $u_{xxx}$ is a known integrable system, while in this paper we study $u_t = u_x^3+u_{xxx}^3+3u_x^2u_{xxx}+3u_{xxx}^2u_x = (u_x+u_{xxx})^3$, leaving the last two PDEs for future researchers to interpret and analyze.}
    \label{fig:sing_val_sols}
\end{figure}

With this procedure, we discovered three PDE families that are a basis for points in the ring of Figure \ref{kdv_messy}. Running CQFinder on all four families of PDEs (including $u_{xxx}$) we retrieve the singular value plots shown in Figure \ref{fig:sing_val_sols}. Future researchers could use L1 regularization as a technique to simplify PDE identification by sparsifying OptPDE's results.

\section{Conserved Quantities of $(u_x+u_{xxx})^3$}

\label{provecq}
As shown in Figure \ref{fig:sing_val_sols}, we discover one non trivial conserved quantity for $u_t = (u_x+u_{xxx})^3$. Here, we verify that this is a real conserved quantity symbolically.

We want to prove $h(u') = 2u_x^2-u_{xx}^2+u_{xx}u$ is a CQ of $f = u_x^3+u_{xxx}^3+3u_x^2u_{xxx}+3u_{xxx}^2u_x$. Namely, that

\begin{equation}
    \frac{dH}{dt} = \int_{-\infty}^{\infty} f \frac{\partial h}{\partial u} + f_x \frac{\partial h}{\partial u_x} + f_{xx} \frac{\partial h}{\partial u_{xx}} = 0.
    \label{eq:totalint}
\end{equation}

Note that $u_x^2+uu_{xx} = \frac{\partial}{\partial x} uu_x$, so we can rewrite $h(u') = u_x^2 - u_{xx}^2$ (removing the trivial part). Then, the derivatives of $h(u')$ are simple: $\frac{\partial h}{\partial u} = 0$, $\frac{\partial h}{\partial u_x} = 2u_x$, and $\frac{\partial h}{\partial u_{xx}} = 2u_{xx}$. The derivatives of $f$ are more complex and will be handled individually. 

The first term in the indefinite integral in Eq. \eqref{eq:totalint} is 0 because $\frac{dh}{du} = 0$.

To calculate the second term in Eq. \eqref{eq:totalint}, we calculate $f_x$ as

\begin{dmath}
    f_x = 3u_x^2u_{xx} + 3u_{xxx}^2u_{xxxx} + 6u_xu_{xx}u_{xxx} + 
    3u_x^2u_{xxxx} + 6u_xu_{xxx}u_{xxxx} + 3u_{xx}u_{xxx}^2,
\end{dmath}

which gives
\begin{dmath}
    f_x \frac{\partial h}{\partial u_x} = 6u_xu_{xxx}^2u_{xxxx} + 12u_x^2u_{xx}u_{xxx} + 6u_x^3u_{xxxx} 
    + 12u_x^2u_{xxx}u_{xxxx} + 6u_xu_{xx}u_{xxx}^2 + 6u_x^3u_{xx}.
\end{dmath}

The third term of Eq. \eqref{eq:totalint} requires $f_{xx}$:

\begin{dmath}
    f_{xx} = 6u_xu_{xx}^2 + 3u_x^2u_{xxx} + 6u_{xxx}u_{xxxx}^2 + 3u_{xxx}^2u_{xxxxx} + 6u_{xx}^2u_{xxx} + 6u_xu_{xxx}^2 + 6u_xu_{xx}u_{xxxx} + 6u_xu_{xx}u_{xxxx} + 3u_x^2u_{xxxxx} + 6u_xu_{xxxx}^2 
 + 6u_xu_{xxx}u_{xxxxx} + 6u_{xx}u_{xxx}u_{xxxx} + 6u_{xx}u_{xxx}u_{xxxx} + 3u_{xxx}^3,
\end{dmath}

which gives 
\begin{dmath}
    f_{xx} \frac{\partial h}{\partial u_{xx}} = -12u_xu_{xx}^3 - 6u_x^2u_{xx}u_{xxx} - 12u_{xx}u_{xxx}u_{xxxx}^2 - 6u_{xx}u_{xxx}^2u_{xxxxx} - 12u_{xx}^3u_{xxx} - 12u_xu_{xx}u_{xxx}^2 - 12u_xu_{xx}^2u_{xxxx} - 12u_xu_{xx}^2u_{xxxx} - 6u_x^2u_{xx}u_{xxxxx} - 12u_xu_{xx}u_{xxxx}^2 - 12u_xu_{xx}u_{xxx}u_{xxxxx} - 12u_{xx}^2u_{xxx}u_{xxxx} - 12u_{xx}^2u_{xxx}u_{xxxx} - 6u_{xx}u_{xxx}^3.
\end{dmath}

Thus, we can express our conservation condition Eq. \eqref{eq:totalint} as

\begin{dmath}
    \label{eq:fullexp}
    \frac{dH}{dt} = \int_{-\infty}^{\infty} f_x\frac{\partial h}{\partial u_{x}} + f_{xx}\frac{\partial h}{\partial u_{xx}} dx
    = 6 \int_{-\infty}^{\infty} \left( u_x^3u_{xx} + u_xu_{xxx}^2u_{xxxx} + u_x^2u_{xx}u_{xxx} + u_x^3u_{xxxx} + 2u_x^2u_{xxx}u_{xxxx} - 2u_xu_{xx}^3 - 2u_{xx}u_{xxx}u_{xxxx}^2 - u_{xx}u_{xxx}^2u_{xxxxx} - 2u_{xx}^3u_{xxx} - u_xu_{xx}u_{xxx}^2 - 4u_xu_{xx}^2u_{xxx} - u_x^2u_{xx}u_{xxxxx} - 2u_xu_{xx}u_{xxxx}^2 - 2u_xu_{xx}u_{xxx}u_{xxxxx} - 4u_{xx}^2u_{xxx}u_{xxxx} - u_{xx}u_{xxx}^3 \right) dx.
\end{dmath}

The value proposition of OptPDE and CQFinder should be  clear when faced with the daunting task of proving that Eq. \eqref{eq:fullexp} is equal to 0. Normally, a physicist would have to propose the integrable system, the conserved quantities, and prove that the quantities are conserved, all of which is aided by our systems. To go about showing that the behemoth Eq. \eqref{eq:fullexp} is 0, we use the transformation $x = ax'$ such that $u_{nx'} = u_{nx} \frac{\partial^n x}{\partial {x'}^n} = a^nu_x$. We can then write Eq. \eqref{eq:fullexp} as 

\begin{dmath}
    \label{eq:splitbya}
    \frac{dH}{dt} = \int_{-\infty}^{\infty} a^5 \left( u_x^3u_{xx} \right) + a^7 \left( u_x^2u_{xx}u_{xxx} + u_{x}^3u_{xxxx} -2u_xu_{xx}^3 \right) + a^9 \left( 2u_x^2u_{xxx}u_{xxxx} -2u_{xx}^3u_{xxx} -u_xu_{xx}u_{xxx}^2 -4u_xu_{xx}^2u_{xxxx} -u_x^2u_{xx}u_{xxxxx} \right) + a^{11} \left( u_xu_{xxx}^2u_{xxxx} -2u_xu_{xx}u_{xxxx}^2 - 2u_xu_{xx}u_{xxx}u_{xxxxx} - 4u_{xx}^2u_{xxx}u_{xxxx} - u_{xx}u_{xxx}^3 \right) + a^{13} \left(-2u_{xx}u_{xxx}u_{xxxx}^2 - u_{xx}u_{xxx}^2u_{xxxxx} \right) dx,
\end{dmath}
where we dropped the factor of 6 from Eq. \eqref{eq:fullexp}. We note that Eq. \eqref{eq:splitbya} should be 0 for all values of $a$. In order for that to be true, the terms associated with each exponent of $a$ must be 0. So, we can split our proof by powers of $a$. 

\subsection{$a^{5}$}

We want to evaluate 

\begin{dmath}
    \int_{-\infty}^{\infty} u_x^3u_{xx} dx.
    \label{eq:a5}
\end{dmath}
If we use $w = u_x$, we can write Eq. \eqref{eq:a5} as 

\begin{dmath}
    \int_{-\infty}^{\infty} w^3 dw = w^4/4 \Big|_{-\infty}^{\infty} = 0,
\end{dmath}
using free boundary conditions. 

\subsection{$a^7$}

We want to evaluate

\begin{dmath}
    \int_{-\infty}^{\infty} u_x^2u_{xx}u_{xxx} + u_{x}^3u_{xxxx} -2u_xu_{xx}^3 dx.
    \label{eq:a7}
\end{dmath}
Note that $\frac{\partial}{\partial x}(u_x^2u_xx^2) = 2u_xu_{xx}^3+2u_x^2u_{xx}u_{xxx}$, so we can rewrite Eq. \eqref{eq:a7} as 

\begin{dmath}
 = u_x^2u_{xx}^2 \Big|_{-\infty}^{\infty} + \int_{-\infty}^{\infty} 3u_x^2u_{xx}u_{xxx}+u_x^3u_{xxxx} = u_x^2u_{xx}^2+u_x^3u_{xxx} \Big|_{-\infty}^{\infty} = 0.
    \label{eq:a7}
\end{dmath}
with free boundary conditions.

\subsection{$a^9$}

We want to evaluate

\begin{dmath}
    \int_{-\infty}^{\infty} 2u_x^2u_{xxx}u_{xxxx} -2u_{xx}^3u_{xxx} -u_xu_{xx}u_{xxx}^2 -4u_xu_{xx}^2u_{xxxx} -u_x^2u_{xx}u_{xxxxx} dx.
    \label{eq:a9}
\end{dmath}
Note that $\frac{\partial}{\partial x} u_x^2u_{xx}u_{xxxx} = 2u_xu_{xx}^2u_{xxxx}+u_x^2u_{xxx}u_{xxxx} + u_x^2u_{xx}u_{xxxxx}$. So we can rewrite Eq. \eqref{eq:a9} as 

\begin{dmath}
    u_x^2u_{xx}u_{xxxx} - \frac{1}{2}u_{xx}^4 \Big|_{-\infty}^{\infty} + \int_{-\infty}^{\infty} 3u_x^2u_{xxx}u_{xxxx} - u_xu_{xx}u_{xxx}^2-2u_xu_{xx}^2u_{xxxx} dx.
    \label{eq:a9r1}
\end{dmath}
Note that $\frac{\partial}{\partial x} 1/2u_x^2u_{xxx}^2 = u_xu_{xx}u_{xxx}^2+u_x^2u_{xxx}u_{xxxx}$. We can rewrite \eqref{eq:a9r1} as  

\begin{dmath}
    u_x^2u_{xx}u_{xxxx} \frac{3}{2}u_x^2u_{xxx}^2 - \frac{1}{2}u_{xx}^4 \Big|_{-\infty}^{\infty} - 2 \int_{-\infty}^{\infty} 2u_xu_{xx}u_{xxx}^2 + u_xu_{xx}^2u_{xxxx} dx.
    \label{eq:a9r2}
\end{dmath}
Note that $\frac{\partial}{\partial x} u_xu_{xx}^2u_{xxx} = \frac{1}{4}\frac{\partial}{\partial x}u_{xx}^4 + 2u_xu_{xx}u_{xxx}^2 + u_xu_{xx}^2u_{xxxx}$. So we can rewrite Eq. \eqref{eq:a9r2} as 

\begin{dmath}
    u_x^2u_{xx}u_{xxxx} + \frac{3}{2}u_x^2u_{xxx}^2  - 2u_xu_{xx}^2u_{xxx} -u_{xx}^4 \Big|_{-\infty}^{\infty}
    = 0. 
    \label{eq:a9r3}
\end{dmath}
with free boundary conditions.

\subsection{$a^{11}$}

We want to evaluate

\begin{dmath}
    \int_{-\infty}^{\infty} \underbrace{u_xu_{xxx}^2u_{xxxx}}_{p_1} - \underbrace{2u_xu_{xx}u_{xxxx}^2}_{p_2} - \underbrace{2u_xu_{xx}u_{xxx}u_{xxxxx}}_{p_3} - \underbrace{4u_{xx}^2u_{xxx}u_{xxxx}}_{p_4} - \underbrace{u_{xx}u_{xxx}^3}_{p_5} \, dx.
    \label{eq:a11}
\end{dmath}

This subintegral was the most difficult to evaluate. We created a computer program to generate all polynomials of $u,u_x, u_{xx}, u_{xxx}, u_{xxxx}, u_{xxxxx}$ up to degree 5 with each of the $p_i$ terms in their first spatial derivative. We retrieve Table \ref{a11table}. We color terms that appear in many $p_i$'s as likely to be in the antiderivative of \eqref{eq:a11}.

Specifically, we propose the antiderivative of Eq. \eqref{eq:a11} to have the form $I_{a^{11}} = c_1u_xu_{xxx}^3+c_2u_{xx}^2u_{xxx}^2+c_3u_xu_{xx}u_{xxx}u_{xxxx}$ for coefficients $c_1, c_2, c_3$. Now, we take the derivative of $I_{a^{11}}$ and match up coefficients with Eq. \eqref{eq:a11} to get the following system of equations: 

\begin{align*} 
3c_1 + c_3 &=  1 \\ 
c_3 &=  -2 \\
c_3 &= -2 \\
2c_2+c_3 &= -4 \\
c_1 + 2c_2 + c_3 &= -1.\\
\end{align*}
The solution to these equations is $c_1 = 1, c_2 = -1, c_3 = -2$. Thus, we can write the integral in Eq. \ref{eq:a11} as 

\begin{dmath}
    u_xu_{xxx}^3 -  u_{xx}^2u_{xxx}^2 - 2u_xu_{xx}u_{xxx}u_{xxxx} \Big|_{-\infty}^{\infty}
    = 0.
\end{dmath}
with free boundary conditions.

\subsection{$a^{13}$}

We want to evaluate

\begin{dmath}
    \int_{-\infty}^{\infty} -2u_{xx}u_{xxx}u_{xxxx}^2 - u_{xx}u_{xxx}^2u_{xxxxx} \, dx.
    \label{eq:a13}
\end{dmath}
We pose the problem in terms of $w = u_{xx}$. Eq. \eqref{eq:a13} becomes

\begin{dmath}
    = -\int_{-\infty}^{\infty} 2ww_xw_{xx}^2 + ww_x^2w_{xxx} \, dx.
    \label{eq:a13r1}
\end{dmath}
Note that we can add $w_x^3w_{xx}$ to Eq. \eqref{eq:a13r1} because it has an integral of 0 over the domain. Eq. \eqref{eq:a13r1} becomes
\begin{dmath}
    = -\int_{-\infty}^{\infty} 2ww_xw_{xx}^2 + ww_x^2w_{xxx} + w_x^3w_{xx}\, dx
    = -ww_x^2w_{xx} \Big|_{-\infty}^{\infty} = 0.
    \label{eq:a13r2}
\end{dmath}
with free boundary conditions.

\subsection{Putting it Together}
Thus, our original Eq. \eqref{eq:totalint} holds, and $u_x^2-u_{xx}^2$ is a conserved quantity of $(u_x+u_{xxx})^3$. What took us three pages of calculus and algebra to show, CQFinder discovered automatically!

\subsection{Numerical Verification}
\begin{figure}
  \includegraphics[width=0.5
  \columnwidth]{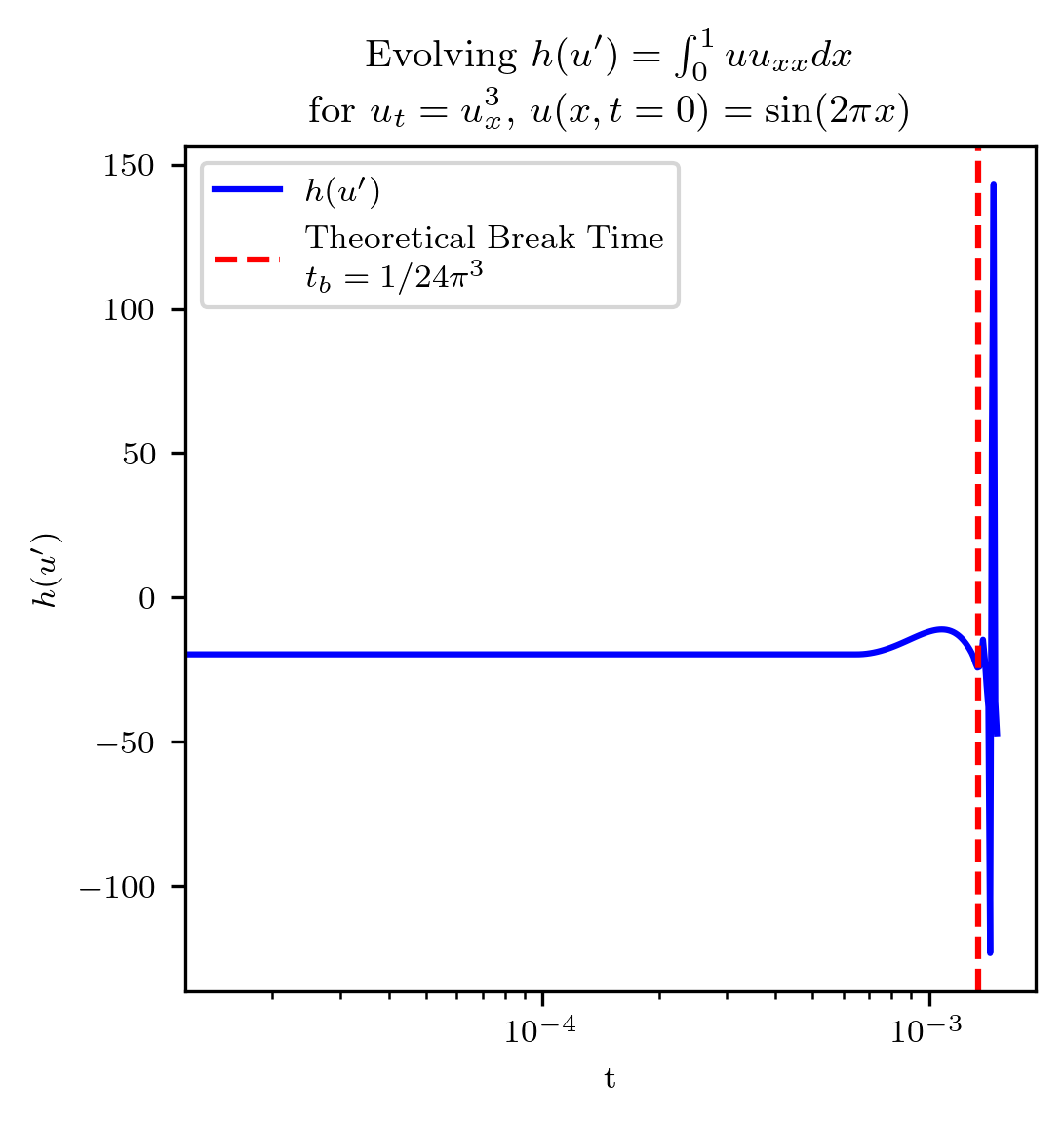}
  \caption{We evolve the CQ $h(u') = uu_{xx}$ for the PDE $u_t = u_x^3$ and see that it is visually conserved until the breaktime of the PDE. Note that these are reduced forms of $h(u') = u_{xx}^2+uu_{xx}$ and $u_t = (u_x+u_{xxx})^3$ transformed with $x = ax'$ for small $a$.}
  \label{fig:evolve_CQ}
\end{figure}

We additionally ensure that our CQ holds for our simulations of the reduced PDE, $u_t = u_x^3$. In this limit, our CQ is $h(u') = \int uu_{xx}$. From Figure \ref{fig:evolve_CQ}, the CQ is visually conserved until the breaktime of the PDE.

\begin{table}[]
\begin{tabular}{@{}|c|l|cccc|@{}}

\toprule
\multicolumn{1}{|l|}{\textbf{$i$}} & \multicolumn{1}{c|}{\textbf{$p_i$}} & \textbf{Terms w/ $p_i$ in Derivative}      &                                            &                                                   &                                                   \\ \midrule
1                                  & $ u_xu_{xxx}^2u_{xxxx}$              & { $ \color[HTML]{3531FF} u_xu_{xxx}^3$}      & $uu_{xxx}^2u_{xxxx}$                       & {\color[HTML]{FE0000} $u_xu_{xx}u_{xxx}u_{xxxx}$} &                                                   \\ \midrule
2                                  & $u_xu_{xx}u_{xxxx}^2$               & $u_x^2u_{xxxx}^2$                          & $uu_{xx}u_{xxxx}^2$                        & {\color[HTML]{FE0000} $u_xu_{xx}u_{xxx}u_{xxxx}$} &                                                   \\ \midrule
3                                  & $u_xu_{xx}u_{xxx}u_{xxxxx}$         & $u_xu_{xx}^2u_{xxxxx}$                     & $u_x^2u_{xxx}u_{xxxxx}$                    & $uu_{xx}u_{xxx}u_{xxxxx}$                         & {\color[HTML]{FE0000} $u_xu_{xx}u_{xxx}u_{xxxx}$} \\ \midrule
4                                  & $u_{xx}^2u_{xxx}u_{xxxx}$           & {\color[HTML]{32CB00} $u_{xx}^2u_{xxx}^2$} & $u_xx^3u_{xxxx}$                           & {\color[HTML]{FE0000} $u_xu_{xx}u_{xxx}u_{xxxx}$} &                                                   \\ \midrule
5                         & $u_{xx}u_{xxx}^2$                   & {\color[HTML]{3531FF} $u_xu_{xxx}^3$}      & {\color[HTML]{32CB00} $u_{xx}^2u_{xxx}^2$} &                                                   &                                                   \\ \bottomrule
\end{tabular}
\caption{Table for the $a^{11}$ integral which shows which terms (right) have derivatives that are terms in Eq. \ref{eq:a11} (left). Highlighted are commonly repeated terms that we hypothesize to be in Eq. \eqref{eq:a11}'s antiderivative.}
\label{a11table}
\end{table}

\section{$u_t=u_x^3$ has an Infinite Number of Conserved Quantities}
\label{sec:infcq}

We assert in the main text that $h(u') = u_x^n$ is conserved for all $n$ for the equation $u_t = u_x^3$. We prove that here. Since $h$ only has $u_x$ terms, we can rewrite the conservation condition, Eq. \ref{eq1} as 

\begin{equation}
    0 = \int_{-\infty}^{\infty} \frac{dh}{du_x}\frac{df}{dx} dx.
    \label{ux3_cons}
\end{equation}
Note that $\frac{dh}{du_x} = nu_x^{n-1}$ and $\frac{df}{dx}=u_x^2u_{xx}$. This allows us to write Eq. \eqref{ux3_cons} as 

\begin{equation}
\begin{split}
    0 &= \int_{-\infty}^{\infty} nu_x^{n-1}u_x^2u_{xx} dx \\
     &= \int_{-\infty}^{\infty} nu_x^{n+1}u_{xx} dx.
\end{split}
\end{equation}

The integral on the right hand side evaluates to $\frac{n}{n+2}u_x^{n+2} \Big|_{-\infty}^{\infty}$ for all $n\neq0,-2$, which evaluates to 0 with free boundary conditions. $n=0$ is trivially conserved because it is constant. $n=-2$ results in the expression $\int_{-\infty}^{\infty} -2\frac{u_{xx}}{u_x} dx = -2\ln{u_x} \Big|_{-\infty}^{\infty}$. With slightly modified ``free" boundary conditions which are symmetric and actually infinitesimally greater than 0, $n=-2$ is also conserved. Thus, $u_x^n$ is conserved for $u_t = u_x^3$ for all $n$, giving $u_t = u_x^3$ an infinite number of conserved quantities and making it an integrable system.

\section{Break Time Mathematical Formulation}
\label{sec:breaktimemath}

We will use the method of characteristics to analyze the $a\ll1$ case of $u_t = (u_x+a^2u_{xxx})^3$, namely $u_t = u_x^3$. 

\begin{figure}
  \includegraphics[width=0.5
  \columnwidth]{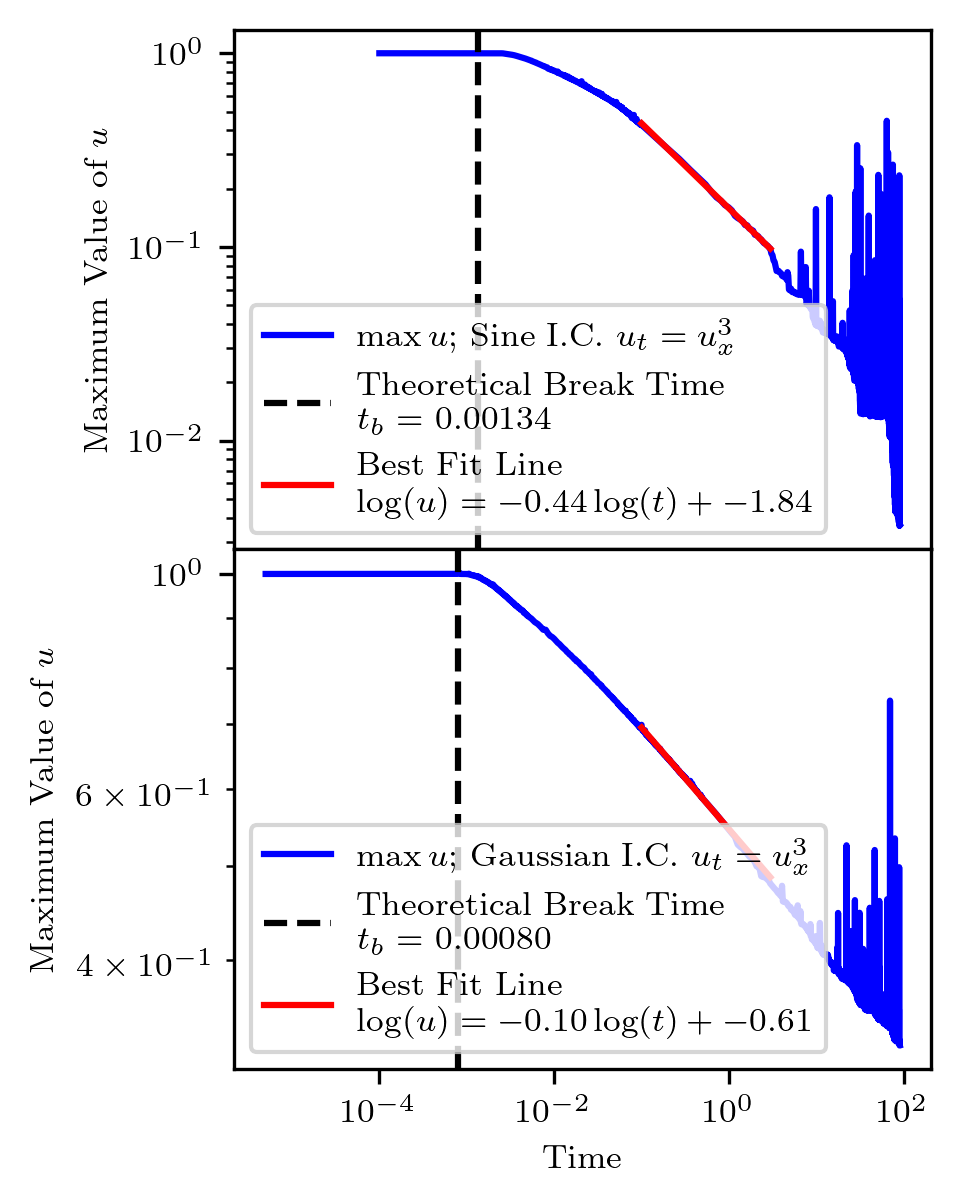}
  \caption{We plot the maximum value of $u$ at every time step in the evolution of $u_t = u_x^3$ for Gaussian and sinusoidal initial conditions. At a certain point, the equation "breaks" and the magnitude of $u$ continues to decay. We calculated this to be the theoretical break time $t_b$ in Appendix \ref{sec:breaktimemath}, which roughly matches up with what we observe. There is a polynomial relationship between time and this magnitude after the break time that is explained by the phenomenological model presented in Section \ref{interp}.}
  \label{maxu}
\end{figure}

First, we note that we can differentiate our equation with respect to $x$ to receive a Burger's like equation: $v_t = 3v^2v_x$ where $v = u_x$, which represents a wave moving with speed $3v^2$. The breaktime of $u_t = u_x^3$ is when this Burger's analogy experiences a shock. 
In the $x-t$ plane, characteristics are lines where $u$ is constant. 
Our characteristics are $\frac{dx}{dt} = 3v^2$ and $\frac{du}{dt} = u_t + u_x\frac{dx}{dt} = u_t +3v^2u_x = 0$. Assuming that the line begins at $x_0$, the equation for the characteristics are $x = 3v^2t+x_0$. The break time is when two sequential characteristics intersect. Since both characteristics have different constant values of $u$, there's a paradoxical shock. The intersection condition can be mathematically formulated as
\begin{equation}
    \begin{split}
        3v(x_0)^2t_b+x_0 &= 3v(x_0+dx)^2t_b+x_0+dx \\
        x_0+dx-x_0 &= 3v(x_0)^2t_b-3v(x_0+dx)^2 \\
        dx &= -t_bd(3v^2) \\
        t_b &= -\frac{1}{\frac{d}{dx}3v^2}.
    \end{split}
\end{equation}
Substituting back $u_x$ for $v$, we find
\begin{equation}
    t_b = \mathrm{argmin}_{t_b > 0} \left\{ -\frac{1}{6u_xu_{xx}} \right\}.
\end{equation}
This roughly lines up with the diminishing magnitude plot, Figure \ref{maxu}.

\section{Phenomenological model derivations}\label{app:phenom}

To understand the wave behavior after breaking, we aim to build a phenomenological model explaining the dynamics when the wave is close to a triangular wave such that the up part and down part can be regarded as linear plus a small quadratic correction $u_{xx}$ ($u_{xx}=C\ll 1$ on the left curve and $u_{xx}=-C$ on the right curve), shown in Figure~\ref{pde_ev} right. 

As described above, each point has a velocity $V=3u_x^2$ moving to the right, while its height does not change. This seems to imply that the wave magnitude will remain a constant, but it actually decays because points are missing at the cusp where two curves meet. Let us denote the point slightly left (right) to the cusp point as $F$ ($S$), whose velocity is denoted as $V_F$ ($V_S$). Since $V_F>V_S$, the left curve will ``eat" into the right curve, causing decrease of the height. At the cusp point, we define the left derivative $(u_x)_F \equiv {\rm tan}\theta_F$ and the right derivative $(u_x)_S \equiv - {\rm tan}\theta_S$. Expressed in terms of $h$ (height), $T$ (half period) and $C$ (second-order derivative), we have $(u_x)_F = \frac{2h}{T} + C\frac{T}{2}, (u_x)_S = \frac{2h}{T} - C\frac{T}{2}, $
So the velocity difference is
$\Delta V\equiv V_F - V_S = 3(u_x)_F^2 - 3(u_x)_S^2 = 12\frac{h}{T}CT = 12Ch$.
Assuming $\theta_1,\theta_2\ll1$, we have $\theta_1\approx\theta_2\approx \frac{2h}{T}$. The derivative of height change is
$\frac{dh}{dt} \approx (V_F-V_S){\rm sin}\theta \approx 24\frac{Ch^2}{T}.$
Further assuming that $C\propto h^\alpha$, we can derive that $h\propto t^{-\frac{1}{1+\alpha}}$. A special case is $\alpha=1$, when the curve is uniformly shrunk along height, we have $h\propto \frac{1}{\sqrt{t}}$, which agrees reasonably well with the sine wave case.


\end{document}